\documentclass{article} 
\usepackage{iclr2020_conference,times}

\usepackage{amsmath,amsfonts,bm}

\def\eqref#1{equation~\ref{#1}}

\def\1{\bm{1}}

\DeclareMathAlphabet{\mathsfit}{\encodingdefault}{\sfdefault}{m}{sl}
\SetMathAlphabet{\mathsfit}{bold}{\encodingdefault}{\sfdefault}{bx}{n}

\usepackage{hyperref}
\usepackage{url}
\usepackage{graphicx}
\usepackage{multirow}

\title{A Neuro-AI Interface for Evaluating Generative Adversarial Networks}

\author{Zhengwei Wang$^1$ \thanks{Work done in the Insight Centre for Data Analytics, Dublin City University.}, Qi She$^2$, Alan F. Smeaton$^3$, Tom\'as E. Ward$^3$ \& Graham Healy$^3$ \\
$^1$ School of Computer Science and Statistics, Trinity College Dublin, Dublin 1, Ireland\\
$^2$ Intel Lab, Beijing, China\\
$^3$ Insight Centre for Data Analytics, Dublin City University, Dublin 9, Ireland\\
\texttt{zhengwei.wang@tcd.ie}, \texttt{qi.she@intel.com}\\
\texttt{\{alan.smeaton,tomas.ward,graham.healy\}@dcu.ie}
}

\iclrfinalcopy

\begin{document}

\maketitle
\vspace{-15pt}
\begin{abstract}
Generative adversarial networks (GANs) are increasingly attracting attention in the computer vision, natural language processing, speech synthesis and similar domains. However, evaluating the performance of GANs is still an open and challenging problem. Existing evaluation metrics primarily measure the dissimilarity between real and generated images using automated statistical methods. They often require large sample sizes for evaluation and do not directly reflect human perception of image quality. In this work, we introduce an evaluation metric called \textbf{Neuroscore}, for evaluating the performance of GANs, that more directly reflects psychoperceptual image quality through the utilization of brain signals. Our results show that Neuroscore has superior performance to the current evaluation metrics in that: (1) It is more consistent with human judgment; (2) The evaluation process needs much smaller numbers of samples; and (3) It is able to rank the quality of images on a per GAN basis. A convolutional neural network (CNN) based \textbf{neuro-AI interface} is proposed to predict Neuroscore from GAN-generated images directly without the need for neural responses. Importantly, we show that including neural responses during the training phase of the network can significantly improve the prediction capability of the proposed model. Codes and data can be referred at this link: \textit{https://github.com/villawang/Neuro-AI-Interface}.
\end{abstract}
\vspace{-15pt}
\section{Introduction}
\vspace{-10pt}
There is a growing interest in studying generative adversarial networks (GANs) in the deep learning community~\citep{goodfellow2014generative}. Specifically, GANs have been widely applied to various domains such as computer vision~\citep{karras2018style}, natural language processing~\citep{fedus2018maskgan}, speech synthesis~\citep{donahue2018synthesizing} and time series generation~\citep{brophy2019quick}. Compared with other deep generative models (e.g. variational autoencoders (VAEs)), GANs are favored for effectively handling sharp estimated density functions, efficiently generating desired samples and eliminating deterministic bias~\citep{wang2019generative}. Due to these properties GANs have successfully contributed to plausible image generation~\citep{karras2018style}, image to image translation~\citep{zhu2017unpaired}, image super-resolution~\citep{ledig2017photo}, image completion~\citep{yu2018generative} etc. 

However, three main challenges still exist currently in the research of GANs: (1) Mode collapse - the model cannot learn the distribution of the full dataset well, which leads to poor generalization ability; (2) Difficult to train - it is non-trivial for discriminator and generator to achieve Nash equilibrium during the training; (3) Hard to evaluate - the evaluation of GANs can be considered as an effort to measure the dissimilarity between real distribution $p_{r}$ and generated distribution $p_{g}$. Unfortunately, the accurate estimation of $p_{r}$ is intractable. Thus, it is challenging to have a good estimation of the correspondence between $p_{r}$ and $p_{g}$. Aspects (1) and (2) are more concerned with computational aspects where much research has been carried out to mitigate these issues~\citep{li2015generative,salimans2016improved,arjovsky2017wasserstein}. Aspect (3) is similarly fundamental, however, limited literature is available and most of the current metrics only focus on measuring the dissimilarity between training and generated images. A more meaningful GANs evaluation metric that is consistent with human perceptions is paramount in helping researchers to further refine and design better GANs.

Although some evaluation metrics, e.g., Inception Score (IS), Kernel Maximum Mean Discrepancy (MMD) and Fr\'echet Inception Distance (FID), have already been proposed~\citep{salimans2016improved,heusel2017gans,borji2018pros}, their limitations are obvious: (1) These metrics do not agree with human perceptual judgments and human rankings of GAN models. A small artifact on images can have a large effect on the decision made by a machine learning system~\citep{koh2017understanding}, whilst the intrinsic image content does not change. In this aspect, we consider human perception to be more robust to adversarial images samples when compared to a machine learning system; (2) These metrics require large sample sizes for evaluation~\citep{empirical-study,salimans2016improved}. Large-scale samples for evaluation sometimes are not realistic in real-world applications since it is time-consuming; and (3) They are not able to rank individual GAN-generated images by their quality i.e., the metrics are generated on a collection of images rather than on a single image basis. 

\citet{yamins2014performance} demonstrates that CNN matched with neural data recorded from inferior temporal cortex~\citep{chelazzi1993neural} has high performance in object recognition tasks. Given the evidence above that a CNN is able to predict the neural response in the brain and can reflect the spatio-temporal neural dynamics in the human brain visual processing area~\citep{cichy2016comparison,tu2018relating,kuzovkin2018activations}, we describe a neuro-AI interface system, where human being's neural response is used as supervised information to help the AI system (CNNs used in this work) solve challenging problems in the real world. As a starting point for exploiting the idea of neuro-AI interface, we focus on utilizing it to solve one of the fundamental problems in GANs: designing a proper evaluation metric.

\begin{figure}[ht!]
	\centering
	\includegraphics[width=.65\textwidth]{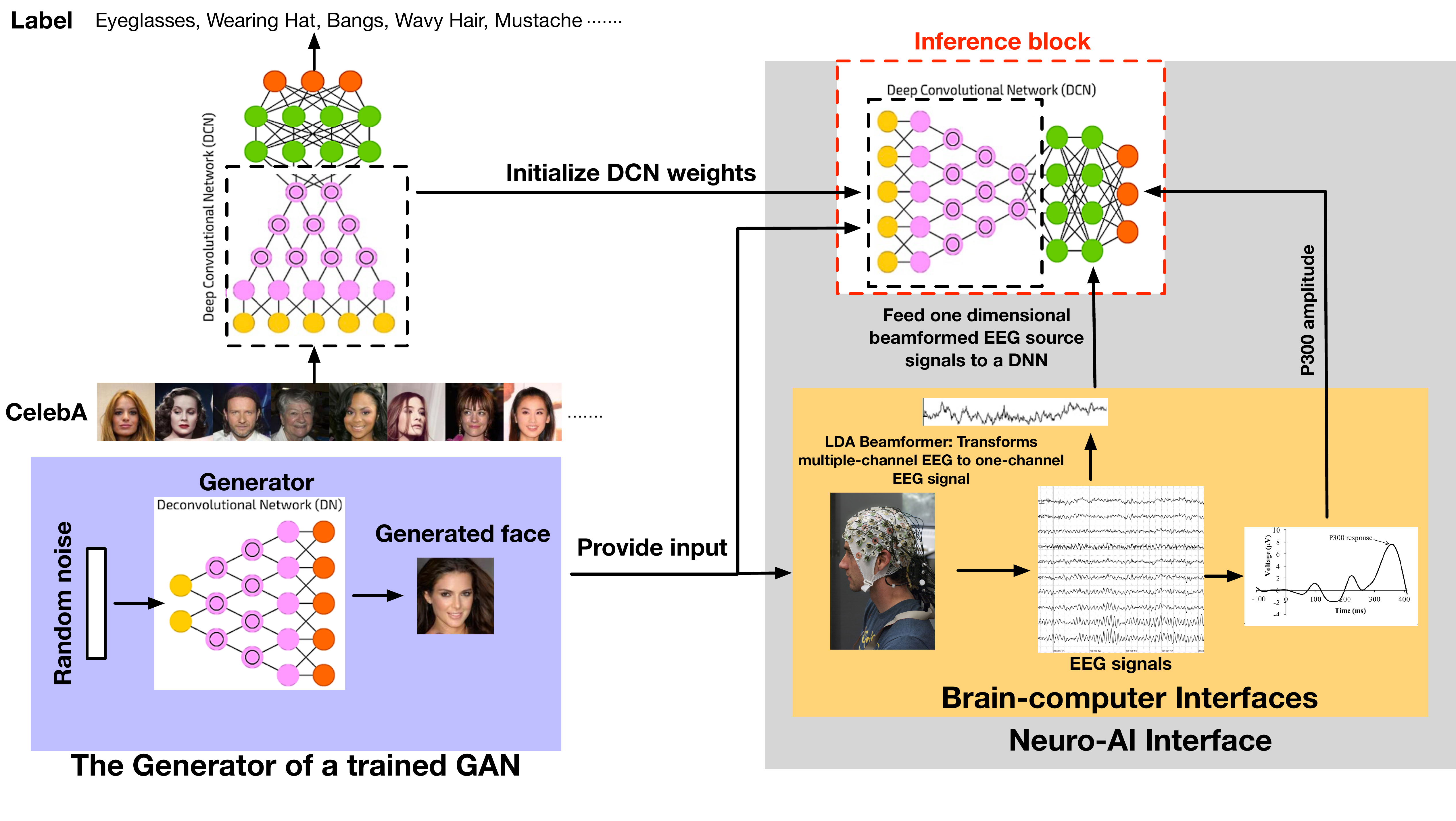}
	\caption{Schematic of neuro-AI interface. Image stimuli generated by GANs are simultaneously presented to a CNN and participants. The inference model is initialized by pretrained weights which has been trained by large scale dataset e.g., CelebA. Participants' P300 amplitude is fed to the network as ground truth and EEG responses are extracted and fed to the CNN as supervisory information for assisting the CNN predict P300 amplitude.}
	\label{fig:neuron_AI_interface1}
\end{figure}

In this paper, we firstly introduce a brain-produced score (Neuroscore), generated from human being's electroencephalography (EEG) signals, in terms of the quality evaluation on GANs. Secondly, we demonstrate and validate a neural-AI interface (as seen in Fig.~\ref{fig:neuron_AI_interface1}), which uses neural responses as supervised information to train a CNN. The trained CNN model is able to predict Neuroscore (we call the predicted Neuroscore as synthetic-Neuroscore) for images without corresponding neural responses. We test this framework via three models: Shallow convolutional neural network, Mobilenet V2~\citep{sandler2018mobilenetv2} and Inception V3~\citep{szegedy2016rethinking}. The scope of the Neuro-AI interface should not be limited in DNN and EEG signals. We think spike trains or fMRI should also be potential source signals that can be used for training AI systems; furthermore, via utilizing more temporal properties~\citep{she2018reduced,she2019neural,feng2019challenges,she2019openloris}, we think the features extracted from these time series signals should be more robust and predictive via learning the temporal structure. 

Neuroscore~\citep{wang2019use} is calculated via measurement of the P300 (by averaging the single-trial P300 amplitude), an event-related potential (ERP)~\citep{polich2007updating} present in EEG, via a rapid serial visual presentation (RSVP) paradigm~\citep{wang2018review,healy2020experiences,wang2019cortically,wang2016investigation,healy2017eeg}. The unique benefit of Neuroscore is that it more directly reflects the perceptual judgment of images, which is intuitively more reliable compared to the conventional metrics~\citep{borji2018pros}. Details of Neuroscore can be refered in~\citep{wang2019use}. 
\vspace{-10pt}

\section{Methodology}
\vspace{-10pt}
\subsection{Neuro-AI Interface}
\begin{figure}[!ht]
	\centering
	\includegraphics[width=.7\textwidth]{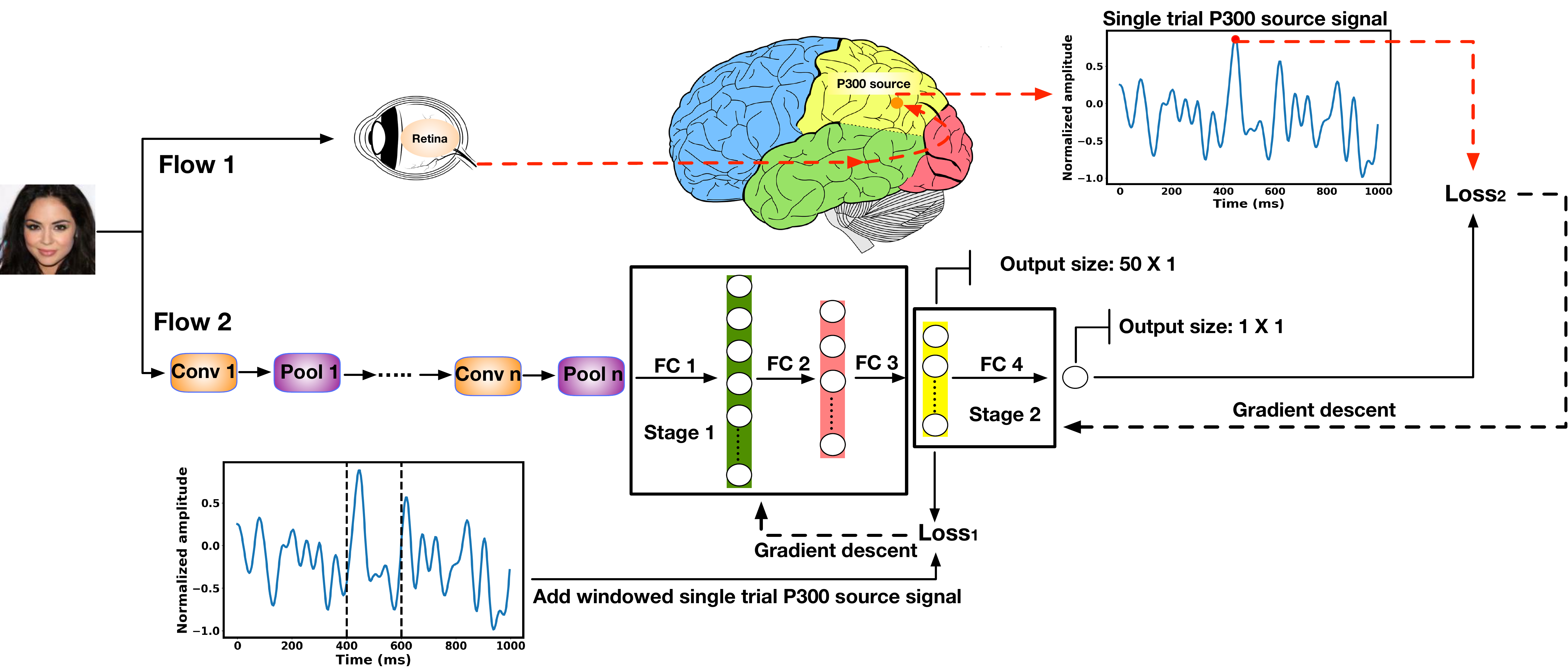}
	\caption{A neuro-AI interface and training details with adding EEG information. Our training strategy includes two stages: (1) Learning from image to P300 source signal; and (2) Learning from P300 source signal to P300 amplitude. $\mathrm{loss}_{1}$ is the $\mathrm{L}_{2}$ distance between the yellow layer and the single trial P300 source signal in the $400$ - $600$ ms corresponding to the single input image. $\mathrm{loss}_{2}$ is the mean square error between model prediction and the single trial P300 amplitude. $\mathrm{loss}_{1}$ and $\mathrm{loss}_{2}$ will be introduced in section~\ref{sec:trianing-details}.}
	\label{fig:inspiration}
\end{figure}

Figure~\ref{fig:inspiration} demonstrates the schematic of neuro-AI interface used in this work.
Flow 1 shows that the image processed by human being's brain and produces single trial P300 source signal for each input image. Flow 2 in Fig.~\ref{fig:inspiration} demonstrates a CNN with including EEG signals during training stage. The convolutional and pooling layers process the image similarly as retina done~\citep{mcintosh2016deep}. Fully connected layers (FC) 1-3 aim to emulate the brain's functionality that produces EEG signal. Yellow dense layer in the architecture aims to predict the single trial P300 source signal in $400$-$600$ ms response from each image input. In order to help model make a more accurate prediction for the single trial P300 amplitude for the output, the single trial P300 source signal in $400$-$600$ ms is fed to the yellow dense layer to learn parameters for the previous layers in the training step. The model was then trained to predict the single trial P300 source amplitude (red point shown in signal trail P300 source signal of Fig.~\ref{fig:inspiration}).

\vspace{-10pt}
\subsection{Training Details} \label{sec:trianing-details}
Mobilenet V2, Inception V3 and Shallow network were explored in this work, where in flow 2 we use these three network bones: such as Conv1-pooling layers. For Mobilenet V2 and Inception V3. We used pretrained parameters from up to the FC 1 shown in Fig.~\ref{fig:inspiration}. We trained parameters from FC 1 to FC 4 for Mobilenet V2 and Inception V3. $\bm{{\mathrm{\theta}}}_{1}$ is used to denote the parameters from FC 1 to FC 3 and $\bm{{\mathrm{\theta}}}_{2}$ indicates the parameters in FC 4. For the Shallow model, we trained all parameters from scratch.

We defined two stage $\mathrm{loss}$ function ($\mathrm{loss}_{1}$ for single trial P300 source signal in the $400$ - $600$ ms time window and $\mathrm{loss}_{2}$ for single trial P300 amplitude) as
\begin{equation}
\begin{aligned}
\mathrm{loss_{1}}(\bm{\mathrm{\theta}}_{1}) &= \frac{1}{N} \sum_{i=1}^{N} \lVert \bm{\mathrm{S}}_{i}^{true} - \bm{\mathrm{S}}_{i}^{pred}(\bm{\mathrm{\theta}}_{1})\rVert ^{2}_{2},\\
\mathrm{loss_{2}}(\bm{\mathrm{\theta}}_{1},\bm{\mathrm{\theta}}_{2}) &= \frac{1}{N} \sum_{i=1}^{N}(\mathrm{y}_{i}^{true} - \mathrm{y}_{i}^{pred}(\bm{\mathrm{\theta}}_{1},\bm{\mathrm{\theta}_{2}}))^{2},
\end{aligned}
\end{equation}
where $\bm{\mathrm{S}}_{i}^{true} \in \mathbb{R}^{1 \times T}$ is the single trial P300 signal in the $400$ - $600$ ms time window to the presented image, and $\mathrm{y}_{i}$ refers to the single trial P300 amplitude to each image.

The training of the models without using EEG is straightforward, models were trained directly to minimize $\mathrm{loss}_{2}(\bm{{\mathrm{\theta}}}_{1},\bm{{\mathrm{\theta}}}_{2})$ by feeding images and the corresponding single trial P300 amplitude. Training with EEG information is visualized in the ``Flow 2" of Fig.~\ref{fig:inspiration} with two stages. Stage 1 learns parameters $\bm{{\mathrm{\theta}}}_{1}$ to predict P300 source signal while stage 2 learns parameters $\bm{\mathrm{\theta}}_{2}$ to predict single trial P300 amplitude with $\bm{{\mathrm{\theta}}}_{1}$ fixed. 

\vspace{-10pt}
\section{Results}
\vspace{-10pt}
Table~\ref{tab:model_loss_cross} 
\begin{table}[!htbp]
	\centering
	\begin{tabular}{c|c|c}
		\hline
		\multicolumn{2}{c|}{Model}                          & Error mean(std)    \\ \hline
		\multirow{3}{*}{Shallow net} & Shallow-EEG           & \textbf{0.209 ($\pm$0.102)} \\ \cline{2-3} 
		& Shallow-EEG$\mathrm{_{random}}$   & 0.348 ($\pm$0.114) \\ \cline{2-3} 
		& Shallow               & 0.360 ($\pm$0.183) \\ \hline
		\multirow{3}{*}{Mobilenet}   & Mobilenet-EEG         & \textbf{0.198 ($\pm$0.087)} \\ \cline{2-3} 
		& Mobilenet-EEG$\mathrm{_{random}}$ & 0.404 ($\pm$0.162) \\ \cline{2-3} 
		& Mobilenet             & 0.366 ($\pm$0.261) \\ \hline
		\multirow{3}{*}{Inception}   & Inception-EEG         & \textbf{0.173 ($\pm$0.069)} \\ \cline{2-3} 
		& Inception-EEG$\mathrm{_{random}}$ & 0.392 ($\pm$0.057) \\ \cline{2-3} 
		& Inception             & 0.344 ($\pm$0.149) \\ \hline
	\end{tabular}
	\caption{Errors of $9$ models for cross participants (``-EEG" indicates models are trained with paired EEG, ``-EEG$\mathrm{_{random}}$" refers to EEG trials which are randomized in the $\mathrm{loss}_{1}$ \textbf{within each type of GAN}). Results are averaged by shuffling training/testing sets for $20$ times. Error is defined as: $\sum_{i}^{m} \lvert \mathrm{Neuroscore}_{pred}^{(i)}  - \mathrm{Neuroscore}_{true}^{(i)} \rvert$, where $\mathrm{m} = 3$ is the number of GAN category used (DCGAN, BEGAN, PROGAN) \citep{radford2015unsupervised,berthelot2017began,karras2017progressive}.}
	\label{tab:model_loss_cross}
\end{table}
shows the error for each model with EEG signal, with randomized EEG signal \textbf{within each type of GAN} and without EEG. All models with EEG perform better than models without EEG, with much smaller errors and variances. Statistic tests between model with EEG and without EEG are also carried out to verify the significance of including EEG information during the training phase. One-way ANOVA tests (P-value) for each model with EEG and without EEG are stated as: $P_{Shallow}=0.003$, $P_{Mobilenet}=0.012$ and $P_{Inception}=5.980e-05$. Results here demonstrate that including EEG during the training stage helps all three CNNs improve the performance on predicting the Neuroscore. The performance of models with EEG is ranked as follows: Inception-EEG, Mobilenet-EEG, and Shallow-EEG, which indicates that deeper neural networks may achieve better performance in this task.

Table~\ref{tab:traditional-method-score} shows the comparison between synthetic-Neuroscore and three traditional scores. To be consistent with all the scores (smaller score indicates better GAN), we used 1/IS and 1/synthetic-Neuroscore for comparisons in Table~\ref{tab:traditional-method-score}. It can be seen that people rank the GAN performance as PROGAN $>$ BEGAN $>$ DCGAN. All three synthetic-Neuroscores produced by the three models with EEG are consistent with human judgment while the other three conventional scores are not (they all indicate that DCGAN outperforms BEGAN).
\begin{table}[ht!]
\small 
\centering
\begin{tabular}{c|c|c|c|c}
\hline
\multicolumn{2}{c|}{Metrics}           & DCGAN & BEGAN & PROGAN \\ \hline
\multicolumn{2}{c|}{1/IS}              &  {0.44} & {0.57} & {0.42}    \\ \hline
\multicolumn{2}{c|}{MMD}               &  {0.22} & {0.29}  & {0.12}      \\ \hline
\multicolumn{2}{c|}{FID}               &  {63.29} & {83.38} & {34.10}       \\ \hline
\multirow{3}{*}{\textbf{Proposed Methods}} & 1/Shallow-EEG   &  {\textbf{1.60}} & {\textbf{1.39}} & {\textbf{1.14}}    \\ \cline{2-5} 
                      & 1/Mobilenet-EEG & {\textbf{1.71}} & {\textbf{1.29}} & {\textbf{1.20}}   \\ \cline{2-5} 
                      & 1/Inception-EEG & {\textbf{1.51}} & {\textbf{1.34}} & {\textbf{1.24}}  \\ \hline
\multicolumn{2}{c|}{Human (BE accuracy)} &{\textbf{0.995}} & {\textbf{0.824}} & {\textbf{0.705}}     \\ \hline
\end{tabular}
\caption{Three conventional scores: IS, MMD, FID, and synthetic-Neuroscore produced by three models with EEG for each GAN category. A lower score indicates better performance of the GAN. Bold text indicates the consistency with human judgment (BE) accuracy.}
	\label{tab:traditional-method-score}
\end{table}

\vspace{-15pt}
\section{Conclusion}
\vspace{-10pt}
In this paper, we introduce a neuro-AI interface that interacts CNNs with neural signals. We demonstrate the use of neuro-AI interface by introducing a challenge in the area of GANs i.e., evaluate the quality of images produced by GANs. Three deep network architectures are explored and the results demonstrate that including neural responses during the training phase of the neuro-AI interface improves its accuracy even when neural measurements are absent when evaluating on the test set. More details of Neuroscore can be found in the recent work~\citep{wang2019neuroscore}.

\bibliography{iclr2020_conference}
\bibliographystyle{iclr2020_conference}


\end{document}